# Combining deep learning and crowdsourcing geo-images to predict housing quality in rural China

Weipan Xu[1], Yu Gu, Yifan Chen, Yongtian Wang, Weihuan Deng, Xun Li

**Abstract**

Housing quality is an essential proxy for regional wealth, security and health. Understanding the distribution of housing quality is crucial for unveiling rural development status and providing political proposals. However, present rural house quality data highly depends on a top-down, time-consuming survey at the national or provincial level but fails to unpack the housing quality at the village level. To fill the gap between accurately depicting rural housing quality conditions and deficient data, we collect massive rural images and invite users to assess their housing quality at scale. As a result, 15,700 rural house images across 28 Chinese provinces are captured. Furthermore, a deep learning framework is proposed to automatically and efficiently predict housing quality based on crowd-sourcing rural images. In conclusion, firstly, the user assessments exceptionally capture the differences in housing quality that are highly related to household income（$R^2$=0.44). Secondly, the machine learning result fits the human result ($R^2$=0.76 at image level and $R^2$=0.90 at the county level), indicating that automatical image assessment is suitable for housing quality. Thirdly, when multiplying per capita housing area by housing quality, we can witness a greater Gini index from 0.29 to 0.31 and map the inequality of rural housing quality at both county and village levels. Our method provides high-resolution estimates of housing quality across China, which could be a complementary tool for monitoring housing change and policy-making and supporting the house-related Sustainable Development Goals.

[1] Email: xuweipan@mail2.sysu.edu.cn
Affiliation: Sun Yat-sen University

# 1 Introduction

Housing quality is paramount for rural residents' wealth. and health. Houses are the most expensive items that most people will consume in their lifetime and are essential contributors to subjective well-being[1–4]. In rural China, housing stock accounts for 31% of the rural wealth, and is regarded as a more accurate barometer of change in the rural income and wealth[5]. Also, housing quality is highly related to human health, such as malaria infection[6,7], child mortality[8], respiratory disease[9]. Correspondingly, Sustainable Development Goal 11 aims for universal access to adequate, safe, affordable housing by 2030[10]. Monitoring the change in the housing quality at a finer-granular scale, especially at the village level will benefit the humanitarian aid and the accomplishment of house-related gaols. However, most researches mainly depend on time-consuming and labor-intensive manual surveys, which is scarce of dynamic and sustainability, especially difficult in remote and undeveloped regions. Crowd-sourcing and machine learning methods are advocated.

Housing quality is defined as the level of housing safety, beauty and habitability. That can be represented by a variation of housing characteristics such as housing size, number of bedrooms, construction materials and housing facilities (e.g., air conditions, indoor toilet, tap water and indoor kitchen)[11,12]. These characteristics could be collected from the household questionnaires such as Demographic and Health Surveys[13] and World Health Surveys[14]. Housing size is the fundamental variable to unpack the economic inequality in rural areas[15,16]. For instance, based on the house area per capita data in China Household Finance Survey, Wang finds the inequality of house assets in rural China is much higher than that in urban China[5]. However, it isn't easy to reflect the whole difference in the house quality if only from the perspective of housing size. Generally, wealthier households can construct more decorated, modernized, high-quality buildings. In contrast, poorer ones can only build basic safe houses, spending less on the appearance, despite of the same housing size. Thus, only measuring house size may underestimate the rural housing inequality. On the contrary, scholars have paid more attention to rural houses' construction materials and facilities[17].

Despite knowing what represents housing quality, how to measure it in a low-cost but more global and finer-granular way is under-researched. Traditional manual surveys are time-consuming, labor-intensive and lack geo-information. To deal with the problem of data scarcity, methods of combining satellite imagery and machine learning are continuously developed. But most of them mainly focus on generating comprehensive indexes for poverty or wealth. From only the perspective of housing quality, Tusting et al. build a spatial regression model to map the housing change in sub-Saharan Africa from 2000 to 2015[17]. House construction materials and house type are the main two predicted housing aspects, and the independent variable includes aridity, urbanicity, accessibility, travel friction, night-time lights and irrigation. But such methods still depend on the quality of the manual surveys and the data availability of independent variables.

On the other hand, crowdsourcing geo-images have been booming recently, providing alternative materials for housing quality evaluation. Geo-images, like street view images, are often used to reflect some region's socio-economic status, like neighborhood income[18], housing price[19], society structure[20], healthy inequality[21], etc. Based on a large-scale dataset of human scoring images, human subjective perception can be learned by machine learning models, especially regarding human feelings of safety, vitality, depression and wealth[22–24]. Unfortunately, most research is carried out in the urban context but not rural.

Estimating rural housing quality from the rural geo-images is a feasible pathway. With social media like TikTok prevailing in rural China, villagers use their mobile phones to record massive rural images and voluntarily share them on social media. This provides a sufficient materials assessment. Geo-images of rural houses can directly reflect the house construction materials, external facilities or even the design, based on which people can judge its quality. For example, multistory, big glass windows, and decorative walls may make people comfortable, while a one-floor house without tiles is less comfortable. Therefore, a robust quality score could be achieved with more people scoring the same house image.

In this paper, we collect massive rural images and invite users to grade housing quality scores according to their subjective assessments. Afterward, we train a deep learning model using these scored images to evaluate house quality according to raw images automatically, efficiently, and at scale. Moreover, based on countrywide evaluation results of rural housing quality, the unequal distribution of housing quality in Rural China is well unveiled at both county and village levels.

## 2 Results

**Housing quality assessment by crowdsourcing**

A total of 15,700 rural house images in 28 provinces across China are collected from a crowdsourcing platform called *Rural Image Shutter,* and 356 volunteers are invited to score the images. Each rural house image is evaluated by at least 15 volunteers. As a result, the housing quality frequency distribution of all house images is displayed in Figure 1b. It could be noted that the average score of rural housing quality is 5.7, while the highest and lowest score is 8.7 and 3.2, respectively. Generally, the housing quality follows the rule of a normal distribution with a standard deviation of 0.85.

Furthermore, we demonstrate some typical rural houses with different quality in Fig.1a. For example, the best-assessed house owns a high score of 0.86 in the left-top corner of Fig.1a. It is characterized by a large area of a homestead. Besides, its facade paving is more delicate with different sizes, textures, and color coordination. Meanwhile, the design of the door building, balcony and roof are well-appointed. In contrast, the worst house in the right-bottom corner of Figure 1a possesses only a tumbled wooden door. The building is embraced by primitive soil walls that seem both insecure and pliable to collapse in a rainstorm.

The indicators of floor height, air conditioning, and external facade can be quantified to demonstrate the main contributor to the housing quality. This paper further divides the houses into groups according to the number of floors, having the air conditioning or not, and the materials of the external facade. As shown in Fig. 1d, the housing quality with one, two, three and above three floors have scores of 5.32, 5.87, 6.16 and 6.20, respectively. The difference in mean values between every two groups pass the student-test, with a *p*-value of 0.001. Similarly, the quality of rural houses with and without air conditioning is 6.00 and 5.61, respectively (Fig. 1e). The housing quality of the ceramic tile, cement, paint and ceramic tile decreased successively, which are 5.88, 7.34, 5.28 and 4.91 points respectively (Fig. 1c). The difference between groups also pass the student-test.

To sum up, we conclude that these objectives affect housing quality with three scales: (a) housing area and floor height. Large houses with more layers will have a relatively high-quality score. (b) external electrical facilities, such as solar water heaters, air conditioners, etc., represent the suitability of the living conditions inside the rural house. (c) façade paving material reflects the mainstream aesthetic of an era and the owner's display of his home equity. Consequently, just owing to these physical objectives, which could be recorded clearly in rural images, people give a quality assessment of rural houses readily and accurately.

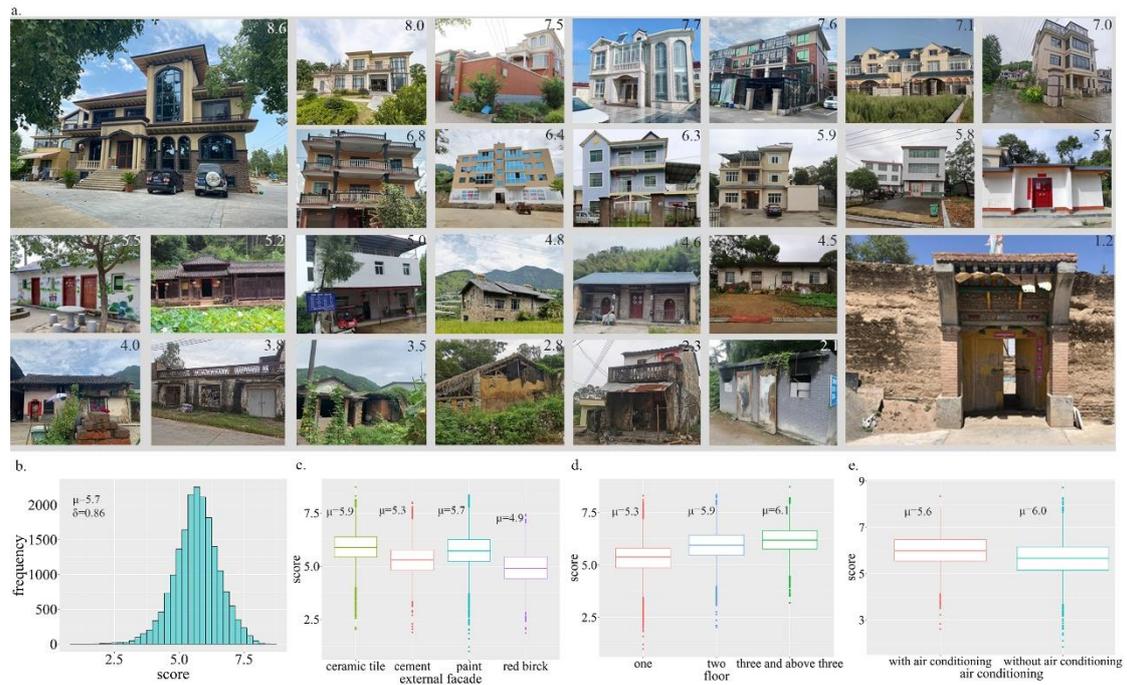

Figure 1 the distribution of housing quality

### Housing quality predicting based on deep learning

Leveraging the massive rural housing images and their quality assessment from *Rural Image Shutter*, we train a deep learning model of DenseNet to predict automatically and efficiently housing quality according to a raw rural house image.

As the result of the model test, the trained DenseNet achieves an $R^2$ of 0.76, a $MSE_{avg}$ of 0.13, and a $MSE_{std}$ of 0.76, respectively. These indicators proved the model has good performance in predicting housing quality with images. In addition, to validate the predicted effectiveness and robustness, we further carry out the housing quality assessment using deep learning in multi-scales. Using machine learning, we are likely to predict the housing quality at any level. Taking all the datasets as input, the relationship is similar to that of the test dataset, with $R^2$=0.87. Scores can be aggregated averagely to village, township and county level. The $R^2$ can reach high values of 0.88 (Fig.2d), 0.90 (Fig.2e) and 0.94 (Fig.2f), respectively.When only considering the test dataset, we can find the r of housing quality also can increase to 0.90 at the county level. These results indicate the model can function well in predicting housing quality from the given images, especially at the county level.

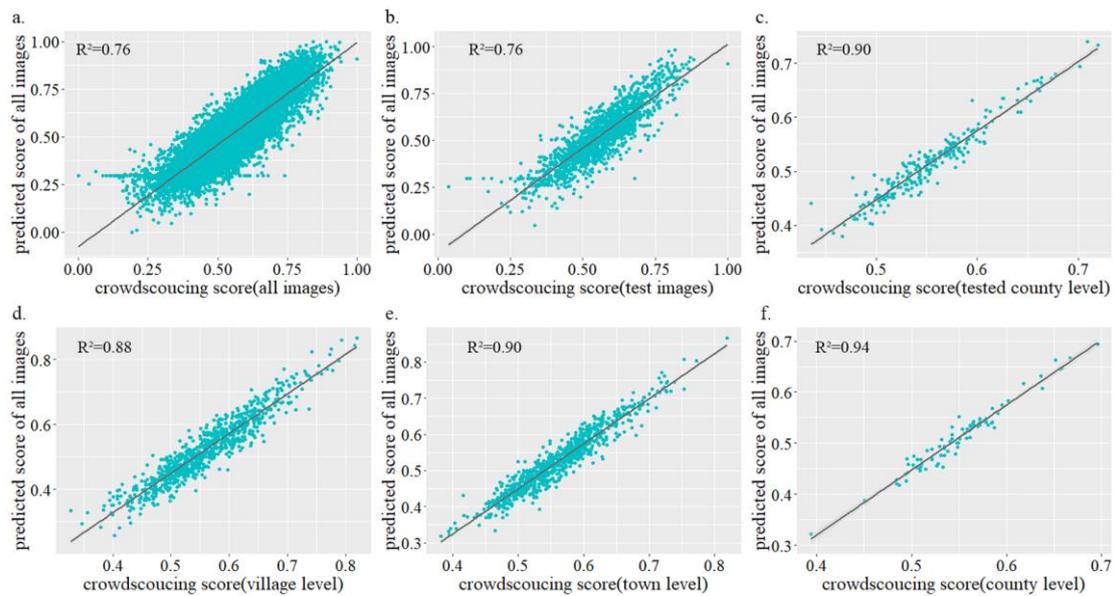

Figure 2 relationship between actual and predicted housing quality.

### Potential to predict large-scale unlabeled images

Another 67,076 unlabeled rural images are collected later and used to test the potential of the deep learning model. *Household Income Index*, rural disposal income per capita and housing area per capita are the three main indicators to check whether the predicted result can match the real rural socio-economic status. As individual economic ability is the important variable affecting the rural house quality, the *Household Income Index*, and rural disposal income per capita are supposed to be highly correlated to the county housing quality. Besides, housing area per capita is more likely to positively affect housing quality.

It could be found that *Household Income Index* is most related to the housing quality with a Pearson's coefficient of 0.64 (Fig.3a). At the same time, the rural disposable income per capita is second related with a Pearson's coefficient of 0.33(Fig.3b). *Household Income Index* is the proportion of households whose annual income is above 60,000 Yuan in a specific county. It is a structured variable that includes some inequality information. In contrast, rural disposable income per capita is a mean variable. The relationship difference indicates there might be much economic inequality between rural individuals. *The household Income Index* can be a better proxy to present the actual rural economic status. As for housing area per capita, Pearson's coefficient reaches a high level of 0.63 (Fig.3c). Therefore, the predicted housing quality can well capture the housing characteristics.

Labeled images are also used to calculate the average county housing quality. It could be found that its relationship to the three economic variables is similar to the result of 67,076 unlabelled images. The Pearson's Coefficients are 0.67(Fig .3d), 0.41(Fig .3e), and 0.61(Fig .3f), respectively. Therefore, we can conclude that our automatic model can predict the housing quality well.

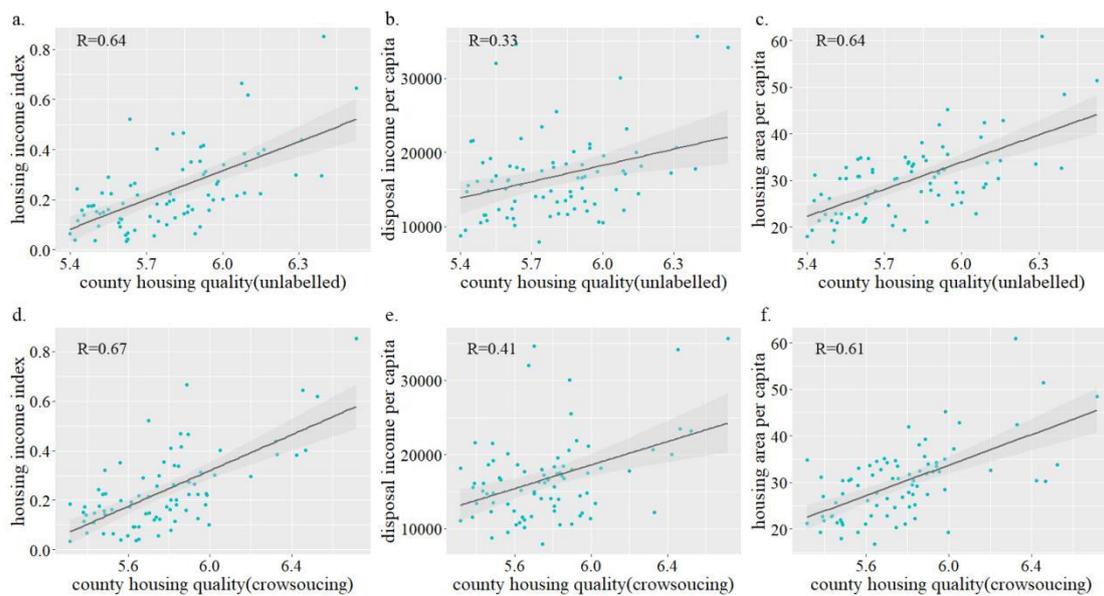

Figure 3 relationship between housing quality and economic variables.

## Inequality of housing quality in rural China

Rural images provide excellent materials for estimating the inequality of housing quality at both village and region levels. From the perspective of regional differences, the quality of rural housing has the differentiation characteristics of "high in the South and low in the north", "high in the East and low in the west"(Fig. 4a). At the village level, the average score of rural houses quality in the southern region is 5.79, higher than that in the northern region of 5.38; The scores of rural housing quality in the eastern, central and western regions progressively decreased, which were 5.69, 5.66 and 5.51 respectively. The north-south gap is greater than the east-west gap, and the score difference between them is 0.41 and 0.18, respectively.

Adding the dimension of rural house quality, the distribution of China's rural housing assets will show a more significant imbalance. The rural housing quality has a strong positive correlation with the rural housing area per capita, with a Pearson's coefficient of 0.63 (Fig. 3c). Therefore, the higher the housing area per capita is, the higher the housing quality is. The housing quality further enhances asset inequality. Specifically, the Gini coefficient of rural residential areas in 81 agricultural counties is 0.29. When considering the weighted housing quality (the product of rural housing quality and per capita living area), the Gini coefficient increases to 0.31, an increase of 7%.

The geo-images visually show the overall wealth level of villages, and the crowdsourcing score further quantifies the level of rural housing assets of each village. Take the villages with more than 5 images as examples. The scores of rural housing quality of 1570 villages are obtained by aggregating the scores of each image. The average score is 5.7, and the villages above the average score account for 48%. The maximum and the minimum values are 7.5 points and 4.0 points, respectively. The village with highest housing quality is located in Jianhu County, Jiangsu province (Fig. 4d). Its rural houses have already reached the modern level, which have harmonious houses with careful design. It mainly follows the traditional architectural style of white walls and black tiles in the south of the Yangtze River. The houses are equipped with complete facilities, and the external wall is hung with air conditioners, water heaters and other equipment. The road hardening, lighting and greening in front of the house are complete. Therefore, in the crowdsourcing scoring process, 219 people and about 22 people have high scores. As for the village with the lowest housing quality, it is located in Jianza county, Qinghai province of undeveloped western China (Fig. 4c). The construction of its rural houses is relatively behindhand, whose walls are still made of adobe. The road in front of the house is dirty and lacks a street lamp. Meanwhile, no other modern facilities are found on the outer wall. For this reason, 114 people and about 23 people scored low. Therefore, the rural geo-images can reflect the imbalance of rural housing quality across rural China.

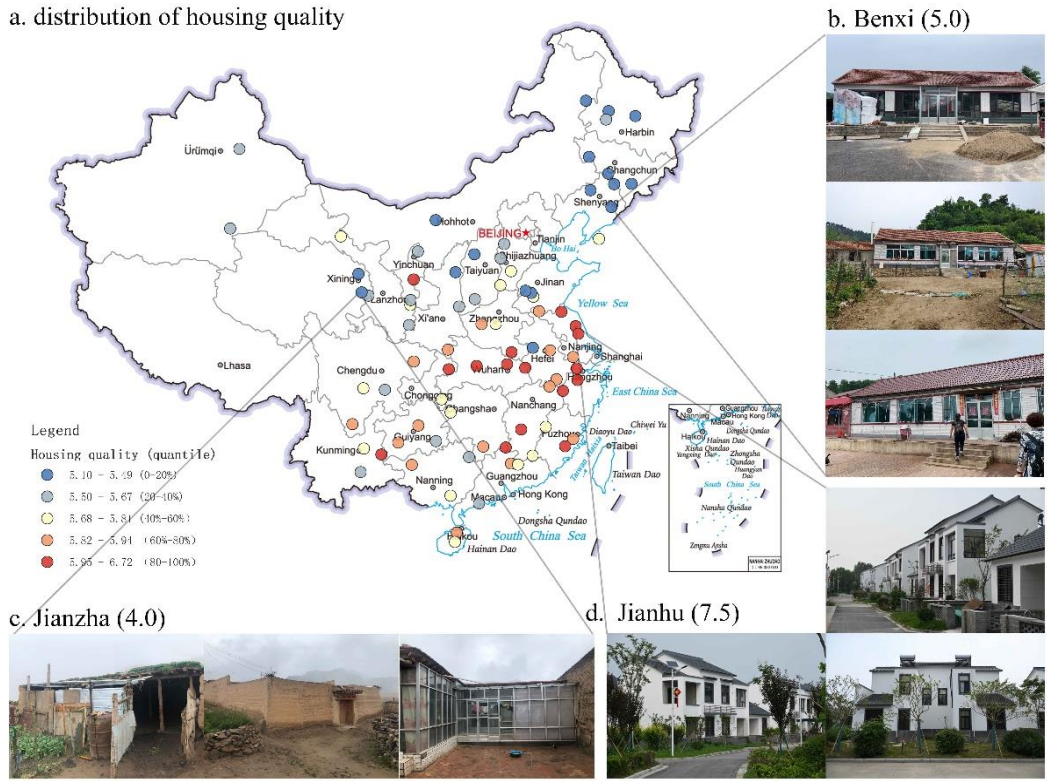

Figure 4. the distribution of county housing quality with some village examples.

## 3 Conclusion and discussion

This study evaluates housing quality across rural China by combining machine learning and crowdsourcing geo-images. In detail, 15,700 rural house images of 28 Chinese provinces are collected from Rural Image Shutter's crowdsourcing platform. The housing quality scoring activity is advocated online, and a total of 345 people join in scoring the rural images. Each image are evaluated by at least 15 people. Based on human-annotated scores, machine learning is trained to promote housing quality estimation work broadly. Thus, inequality of housing quality can be unpacked across China at both village and county levels, which can contribute to precise humanitarian aid.

To conclude, firstly, crowdsourcing score activities can well capture the housing quality from the geo-images. Housing quality can be depicted by a set of tangible characteristics, like windows, construction materials, design style, etc. Those indicators can easily be observed by people through images. In quantitative, more floors, air conditioners, and decorations mean better housing quality. Housing quality is also highly related to household income, with a Pearson's coefficient of 0.66. Secondly, the machine learning result can highly fit the result of human scoring. As for the 1576 tested images, the Pearson's coefficients of real and predicted scores are 0.87 at the image level and 0.95 at county level. The model can be generalized to more similar work, reducing the cost of labor and time. Finally, the distribution of housing quality is unbalanced, with southern China outperforming northern China, and eastern China surpassing western China. Compared with only considering the per capita residential area, the

Chinese rural asset expresses a more serious inequity of an increasing Gini coefficient of 7%, when weighted with house quality and per capita residential area.

This paper advocates crowdsourcing to solve the problem of collecting rural dispersive images. Crowdsourcing scoring on the rural image is a feasible way to get the true housing quality. Thanks to the popularity of WeChat, Tiktok in all aspects of rural daily life, rural areas can generate socio-economic images from the bottom up rather than by top-town survey methods. In other words, rural area data is no longer collected passively but produced actively. Images include high dimensions but visual socio-economic information. As for housing quality, it is easy for rural residents to distinguish high-quality houses from low-quality ones. With more and more people attending the scoring activities, we may get a robust result and predict the direction in the change of rural residents' pursuit of modernized houses. In addition, geo-images are usually attached with detailed geographic information, making it possible to aggregate the housing characteristic to higher levels, such as village and county levels. Therefore, we are more likely to analyze the spatial inequality of housing quality at any level, contributing to a more precise humanitarian aid. The machine learning method is another contribution of this paper. With our model, it will take less time and cost to evaluate the housing quality from the rural geo-image and build up a dynamic monitoring system. To optimize the model, all annotated data and models are open-access.

## 4 Methods

**Crowd-sourcing rural image uploading and marking based on *Rural Image Shutter*.**

*Rural Image Shutter* is a crowd-sourcing platform, providing an online application for users to share their daily life by uploading rural images. In broad rural areas in China, villagers record a variation of elements around themselves such as houses, farmland, ponds, gardens, roads, etc., that bring the emergent potential for understanding rural development and status. This study focuses on leveraging rural images of houses to evaluate the housing quality automatically at a national scale and further reveal the inequality of rural house distribution.

First, rural house images from *Rural Image Shutter* are filtered as the dataset. Specifically, 15,700 housing images are captured covering 83 counties in 28 provinces, China (Fig. 5a). These images distribute widely and represent the holistic styles and features of China's rural house. Besides, the geotagged information of these images provides a possibility to discover their spatial variation.

Moreover, *Rural Image Shutter* invites numerous users to evaluate the housing quality according to the captured house images(Fig. 5b). In reality, humans visually evaluate the rural housing quality by observing the physical elements, for instance, the construction materials, building height, decoration style, and household facilities. In conclusion, all rural house images are evaluated by at least 15 users, and all images are graded from 0 to 10; a higher score indicates better quality.

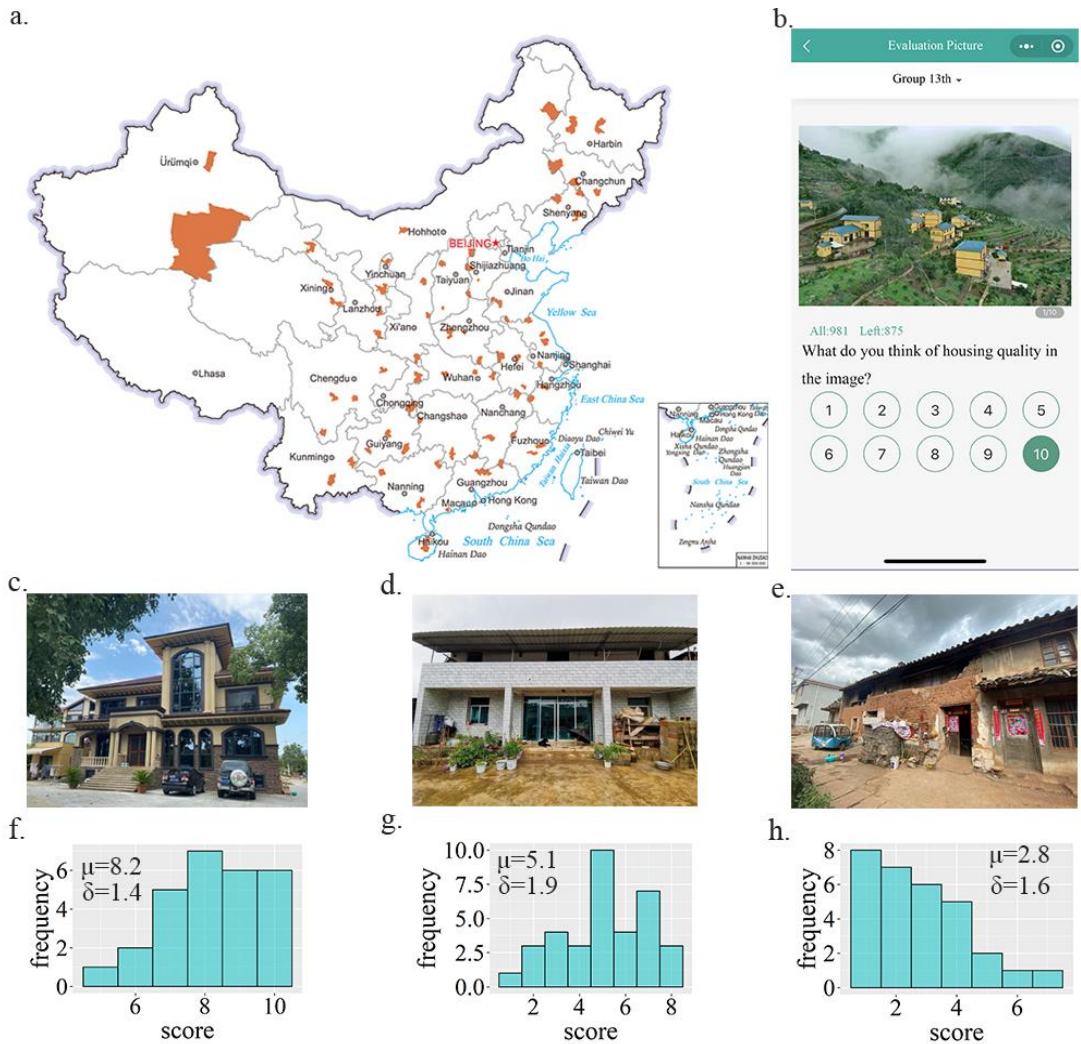

Figure 5 Crowd-sourcing *Rural Image Shutter* platform. a) spatial distribution of rural house image across 83 Chinese counties. b) the interface of crowdsourcing scoring for housing quaility. c-h) rural housing with different quality from high, medium to low quality, respectively and corresponding scores

To finish the above housing quality evaluation work, a crowdsourcing scoring platform is developed based on the WeChat applet for mobile phones (Fig 5b). Numerous network citizens can easily access it. The following evaluation criteria are formulated for the quality of rural houses, including 1) Whether the wall of the house is aesthetic and practical; 2) Whether the roof structure is aesthetic or a local specialty; 3) Whether the style of doors and windows is aesthetic; 4) Whether contain balconies, steps, canopies, ramps, chimneys and other auxiliary parts of the house; 5) Whether have air conditioning, parking spots, and other facilities; 6) Whether have a strong desire to live, etc. Take three images of different quality level as examples. Fig 5c, 5d and 5e represent high, medium and low quality, with the mean scores of 8.2, 5.1 and 2.8, respectively. Also, the histogram of housing quality of each image could be seen in Fig. 5f, Fig. 5g and Fig. 5i. The controversy can be seen by the standard deviation. Afterward, house images in different villages are aggregated into 14 groups, each containing more than 1,000 images. Finally, the mean value of the housing quality score of each image is used as the final result for the follow-up study and analysis. Moreover, the standard deviation can be used as the evaluation standard to reflect the differences in the level of experts. Considering that the different regions where experts live may affect the scoring results, experts from various regions of the country are invited to score the images of rural houses on the scoring platform to reduce regional differences in assessments. Moreover, the standard deviation can be used as the evaluation standard to reflect the differences in the level of experts.

**Deep learning for housing quality assessment**

As the before-mentioned description, based on *Rural Image Shutter* platform, we derived massive geotagged rural house images and their corresponding quality scores. Accordingly, it provides the opportunity for automatically and at scale assessing house quality with rural house images using deep learning

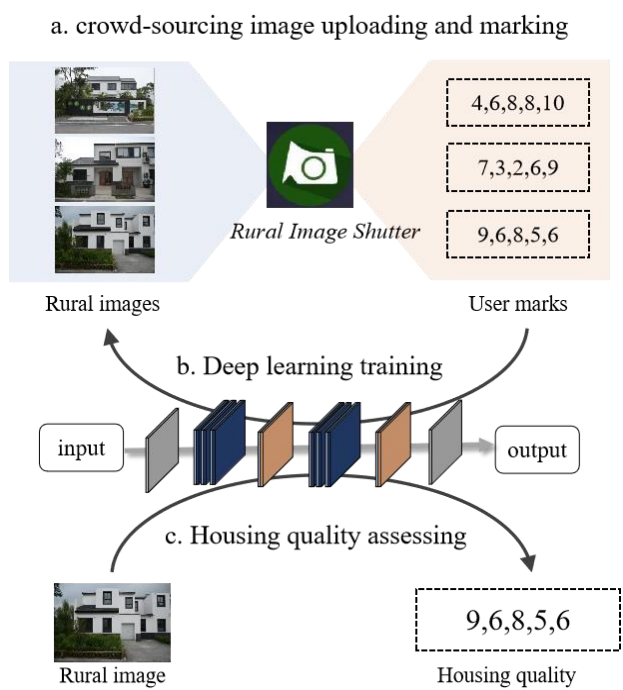

Figure 6 Framework of assessing housing quality with house images using deep learning

The framework of assessing housing quality with house images using deep learning is demonstrated in Fig.6. (A) the Rural Image Shutter platform collects massive rural house images and quality scores. (B) these images and their scores are used to train a deep learning model. (C) when a user uploads a raw rural house image, its corresponding house quality score will be predicted immediately.

To accurately and efficiently predict housing quality with rural house images, DenseNet is adopted as the deep learning model. Compared with traditional Deep Neural Network like VGG and ResNet, DenseNet has good capacity to extract image feature and finally get striking success in a series of image recognition task(Fig .7). Therefore, it is suitable for it our housing quality predicting with massive images.

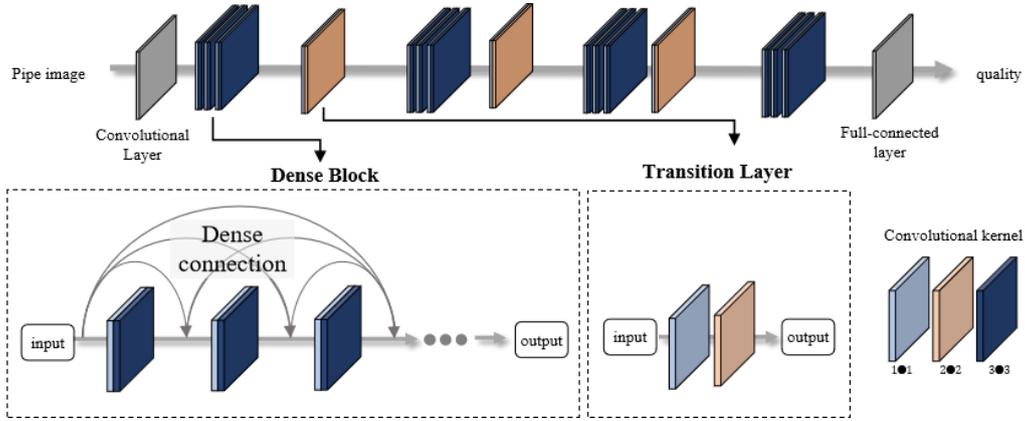

Figure 7 The architecture of DenseNet

The architecture of DenseNet is shown in Fig.7, it consists of a single convolutional layer, four Dense Blocks, three Transition layers and one full-connected layer in sequence. To be specific, as shown in the figure, the Dense Block is comprised of several modules of two different convolutional kernels. Furthermore, each output feature will be connected with latter features, namely "dense connection", which fully takes advantage of features from shallow layers to enhance model performance. Otherwise, the Transition layer is used to connect the adjacent Dense Blocks. In summary, a housing image is input and constantly convolved to be high-dimensional features, then the full-connected layer predicts the housing quality based on the features.

**Model training**

For Deep Neural Network, the predicted accuracy is determined by the parameters of models; therefore, in order to obtain the optimal value of parameters, the derived massive house images and their quality scores are used to train the model to update parameters which is called Back Propagation until the loss function achieves convergence. In model training, we use loss functions of Mean Square Error (MSE) to measure the deviation between predicted and true house quality in each iteration, formulated in Eq.1.

$$MSE = \frac{\sum_{j=1}^{10} \frac{\sum_{i=1}^{n}(\hat{y}_i^j - y_i^j)^2}{n}}{10} \quad (1)$$

Where, $\hat{y}_i^j$ and $y_i^j$ is respectively the predicted and true possibility of score $j (j \in [1,10])$ of image $i (i \in [1, n])$.

**Experimental set**

The total of 15,700 rural house images and their possibilities of 10 quality levels are as dataset and divided into training, validation and test set with the proportions of 80%, 20%, and 10% respectively. Otherwise, some hyper-parameters of DenseNet model is set as below: batchsize is 32, epochs is 100, and learning rate is initially $1\times 10^{-5}$ and adaptively adjusts with the decreasing degree of 0.1.

**Model performance evaluation**

After model training, the performance of the trained DenseNet is evaluated by $R^2$ $MSE_{avg}$, and $MSE_{std}$ formulated in Eq.2 to Eq.4.

$$MSE_{avg} = \frac{\sum_{i=1}^{n}(\hat{y}_i^{avg}-y_i^{avg})^2}{n} \quad (2)$$

$$MSE_{std} = \frac{\sum_{i=1}^{n}(\hat{y}_i^{std}-y_i^{std})^2}{n} \quad (3)$$

$$R^2 = 1 - \frac{\sum_{i=1}^{n}(\hat{y}_i^{avg}-y_i^{avg})^2}{\sum_{i=1}^{n}(\bar{y}^{avg}-y_i^{avg})^2} \quad (4)$$

Where, $\hat{y}_i^{avg}$ and $y_i^{avg}$ are the predicted and true weighted average of quality scores of images $i$, respectively; $\hat{y}_i^{std}$ and $\hat{y}_i^{std}$ are the predicted and true standard deviation of quality scores of images $i$; $\bar{y}^{avg}$ is the average of $y_i^{avg}$.

In detail, $R^2$ can examine the fitting degree between dependent and independent variables of the model, where the result of 1 demonstrates a perfect fit, and it means a reliable model for predictions. $MSE_{avg}$ elucidates the average deviations between an average of predicted and true value while $MSE_{std}$ represent the similar implication replaced by standard deviation. The two MSE indicators could reflect the accuracy between predicted and true quality possibility of house images.